\documentclass[runningheads]{llncs}
\usepackage{graphicx}
\usepackage{cite}
\usepackage{amsmath,amssymb,amsfonts}
\usepackage{algorithmic}
\usepackage{graphicx}
\usepackage{textcomp}
\usepackage{xcolor}
\usepackage{multirow}
\usepackage{subcaption}
\usepackage{todonotes}

\usepackage{hyperref}

\def\BibTeX{{\rm B\kern-.05em{\sc i\kern-.025em b}\kern-.08em
    T\kern-.1667em\lower.7ex\hbox{E}\kern-.125emX}}

\newcommand{\norm}[1]{\left\lVert#1\right\rVert}

\newcommand{\ttilde}{{\raise.17ex\hbox{$\scriptstyle\sim$}} }

\begin{document}

\title{Inversion of Time-Lapse Surface Gravity Data for Detection of 3D CO$_2$ Plumes via Deep Learning}

\titlerunning{Time-Lapse Gravity Inversion via Deep Learning}

\author{Adrian Celaya\inst{1} \and
Bertrand Denel \inst{2} \and
Yen Sun \inst{1} \and
Mauricio Araya-Polo\inst{1} \and
Antony Price \inst{2}}
\authorrunning{A. Celaya et al.}

\institute{TEEP Research and Technology USA, Houston, TX 77002, USA \and
TotalEnergies, France}
\maketitle              % typeset the header of the contribution
\begin{abstract}
We introduce three algorithms that invert simulated gravity data to 3D subsurface rock/flow properties. The first algorithm is a data-driven, deep learning-based approach, the second mixes a deep learning approach with physical modeling into a single workflow, and the third considers the time dependence of surface gravity monitoring. The target application of these proposed algorithms is the prediction of subsurface CO$_2$ plumes as a complementary tool for monitoring CO$_2$ sequestration deployments. Each proposed algorithm outperforms traditional inversion methods and produces high-resolution, 3D subsurface reconstructions in near real-time. Our proposed methods achieve Dice scores of up to 0.8 for predicted plume geometry and near perfect data misfit in terms of $\mu$Gals. These results indicate that combining 4D surface gravity monitoring with deep learning techniques represents a low-cost, rapid, and non-intrusive method for monitoring CO$_2$ storage sites.

\end{abstract}

\section{Introduction}
Greenhouse Gas (GHG) reduction commitments aim to address the increase of atmospheric CO$_2$ concentration implemented on a large scale. They include technologies such as fossil fuel consumption reduction (improvement of energy efficiency, application of new energy sources), expansion of absorption sources through afforestation/reforestation, and carbon dioxide capture and storage (CCS).

One of the CCS technologies currently being deployed worldwide is CO$_2$ geological storage. The technology involves separating and capturing CO$_2$ emitted from large-scale fixed sources, such as cement, petrochemical or coal-fired power plants, and storing it in saline formations located deep underground (hereinafter referred to as "aquifers"). It holds promise as the most practical, immediate-effect, and near-term technology because a range of technologies accumulated in fields such as oil-well drilling, underground storage of natural gas, and Enhanced Oil Recovery (EOR) are readily adaptable.

Ongoing monitoring of these geological storage sites is mandated by regulatory authorities and will need to demonstrate volume storage integrity over long periods. A passive measure of density distribution changes in the subsurface, 4D gravity monitoring is a technique of interest as it is low-cost, rapid, and has a minimal environmental impact. Indeed, 4D gravity monitoring techniques have been deployed to aid hydrocarbon production monitoring in the Norwegian and North Sea sectors for approximately the last 20 years. 

\subsection{Literature Review}
Conventional inversion methods find a model that has the minimum possible structure and whose gravity response fits the observed data \cite{degrood1990, li1998, nabighian2005}. The minimum structure is achieved by minimizing model roughness through a non-linear least squares regression, resulting in a smooth model. While the least squares regression produces smooth density models, the predicted models are often larger and exhibit smaller density values than the actual model \cite{boulanger2001, rezaie2017}. 

Deep learning (DL) is an emerging alternative to traditional geophysical inversion \cite{jin2017, araya2018, kim2018, yang2019}. Over the last several years, deep convolutional neural networks (CNNs) have have achieved state-of-the-art results in a variety of computer vision applications such as image classification, segmentation, and generation \cite{imagenet, brats2, stylegan}. CNNs have recently been used for inversion of seismic \cite{adler21, chen2020, li2020}, electromagnetic imaging \cite{colombo2020, oh2020}, and electrical resistivity data \cite{liu2020, shahriari2020}. Concerning gravity data, Q. Yang et al. use a 3D U-Net to invert surface gravity data to recover synthetic subsurface density anomalies \cite{yang2022}. Their results demonstrate that DL is an effective approach for inverting 2D surface gravity data. However, the density anomalies used in this study are 3D rectangular prisms with constant density values. These anomalies are unrealistic, and their methods are not tested on realistic-looking data. Wang et al. use a 3D U-Net++ to invert more complex density anomalies (in terms of geometry) and apply their method to the inversion of the San Nicolas mining area. While these results are promising, they are not tested on data from CO$_2$ storage sites \cite{wang2022}.

X. Yang et al. proposed a U-Net architecture for the inversion of surface gravity for monitoring CO$_2$ plumes \cite{plume-detection}. Their model successfully detected various synthetic CO$_2$ plumes in most test scenarios. However, the CO$_2$ plumes used in this study were unrealistic regarding model shape and saturation distribution. Additionally, their proposed method maps 2D surface gravity maps to 2D cross-sections of the 3D CO$_2$ plumes they generate. While this study is a promising proof of concept, it does not test DL-based inversion of surface gravity data in a realistic setting. Um et al., present a more realistic test of DL-based inversion for CO$_2$ plume monitoring \cite{um2022}. Using simulated CO$_2$ plumes from the Kimberlina site, they developed a 2D DL architecture to perform joint inversion with seismic, electromagnetic, and gravity data. Additionally, they use a modified version of their architecture to invert their imaging modalities individually. In each case, their DL-based approach can recover CO$_2$ plumes. However, their approach still does not perform DL-based inversion in a fully 3D setting. Alyousuf et al. perform 3D inversion on realistic, physics simulated CO$_2$ plumes (Johansen formation) using three-axis borehole data as the input to their proposed architecture \cite{alyousuf2022}. Their results indicate that DL based-methods can successfully invert borehole gravity data. However, borehole data is highly localized and does not give a global picture of the storage site. 

\subsection{Novel Contributions}
This paper makes the following novel contributions:
\begin{enumerate}
    \item We develop an effective 3D DL-based inversion method to recover high-resolution subsurface CO$_2$ plumes from surface gravity data. To the best of our knowledge, this is the first fully 3D approach for the inversion of surface gravity data which is tested on realistic, physics-simulated CO$_2$ plumes.
    \item We develop a postprocessing method that combines the strengths of traditional and DL-based inversion. 
    \item We develop a hybrid, physics-informed workflow for DL-based inversion.
    \item We develop a time-dependent approach for DL-based inversion of surface gravity data. To the best of our knowledge, this is the first DL-based approach to consider time dependence for inverting surface gravity data.
    \item We explore the effects of different sensor resolution grids to DL-based inversion.
    \item We demonstrate the ability of DL-based methods to be easily adapted to different geophysical targets like density and saturation.
\end{enumerate}

The paper is organized as follows: Section~\ref{sec:data} reviews how the data is prepared. Section~\ref{sec:methods} describes the methods, from pure data-driven to a physics-driven one.
Sections~\ref{sec:eval} discusses experimental setup and results. In section Discussion~\ref{sec:diss} perspective and analysis is provided.
Section \ref{sec:conclu} introduce the conclusions and road map for the proposed approach.

\section{Data Preparation}
\label{sec:data}
% \todo[inline]{do we need to mention the fully synth models initially used (like the LANL folks)? what about Snohvit?}
Located 60km offshore the west coast of Norway, the Johansen formation is a candidate CO$_2$ storage site whose theoretical capacity exceeds 1Gt CO$_2$. It consists of a roughly 100 m thick aquifer that extends up to 100 km in the north-south direction and is 60 km wide. The depth ranges from 2200 to 3100m below sea level, which offers the perfect conditions of pressure and temperature to inject CO$_2$ in a supercritical state.

As part of the MatMoRA project, a corresponding reservoir model is made available online by SINTEF \cite{eigestad}. Starting from this initial setup, we use geostatistics to generate 500 new geological realizations that all differ in terms of porosity and permeability. In this work, we relied on a Fast Gaussian procedure (as detailed in \cite{lorentzen}), whose main parameters are given in Table \ref{tab:geostats}.

% \begin{table}[h!]
%   \begin{center}
%     \begin{tabular}{c|c|c}
%       \multirow{2}{*}{Porosity} & Std & 0.03\\
%       & Lower bound & 0.1\\ 
%       & Upper bound & 0.4\\ 
%       \hline
%       \multirow{2}{*}{Permeability (log)} & Std & 2\\ 
%       & Lower bound & -5\\
%       & Upper bound & 10\\ 
%       \hline      
%       \multirow{2}{*}{Correlation length} & Mean & 26\\ 
%       & Std & 2\\
%       \hline      
%       Correlation & Porosity-Permeability & 0.3      
%     \end{tabular}
%     \caption{Parameters for geostatistics simulation of Johansen formation. \label{tab:geostats}}
% \end{center}
% \end{table}

% \usepackage{multirow}
% \usepackage{graphicx}
\setlength\tabcolsep{10pt}
\begin{table}[ht!]
\centering
\bgroup
\def\arraystretch{1.25}%  1 is the default, change whatever you need
\begin{tabular}{llc}
\hline
\multirow{3}{*}{Porosity}           & Std                   & 0.03 \\
                                    & Lower Bound           & 0.10 \\
                                    & Upper Bound           & 0.40 \\ \hline
\multirow{3}{*}{Permeability (log)} & Std                   & 2.00 \\
                                    & Lower Bound           & -5.0 \\
                                    & Upper Bound           & 10.0 \\ \hline
\multirow{2}{*}{Correlation Length} & Mean                  & 26.0 \\
                                    & Std                   & 2.00 \\ \hline
Correlation                         & Porosity-Permeability & 0.30 \\ \hline
\end{tabular}%
\egroup
\caption{Parameters for geostatistics simulation of Johansen formation.}
\label{tab:geostats}
\end{table}

For each realization, we carry out a fluid flow simulation to model the evolution of the CO$_2$ saturation plume within the reservoir. The scenario consists of a 100-year-period of CO$_2$ injection at a constant rate of 14400 $\text{m}^3/\text{day}$, followed by a 400-year-period of migration. We perform these simulations using the MRST toolbox \cite{lie_2019}, which implements a Vertical Equilibrium model.

Assuming that the only changes in density occur within the reservoir (with CO$_2$ replacing brine), one can then write the time-lapse bulk density evolution as:
\begin{equation}
    \Delta \rho = \phi  \Delta S_{CO_2} (\rho_{CO_2} - \rho_{brine})
\end{equation}
where $\phi$ is the porosity, $\Delta S_{CO_2}$ is the change in CO$_2$ saturation, $\rho_{CO_2}$ and $\rho_{brine}$ stand for the CO$_2$ and brine density, respectively (see \cite{gasperikova}).

\begin{figure}[ht!]
    \centering
    \includegraphics[width = \textwidth]{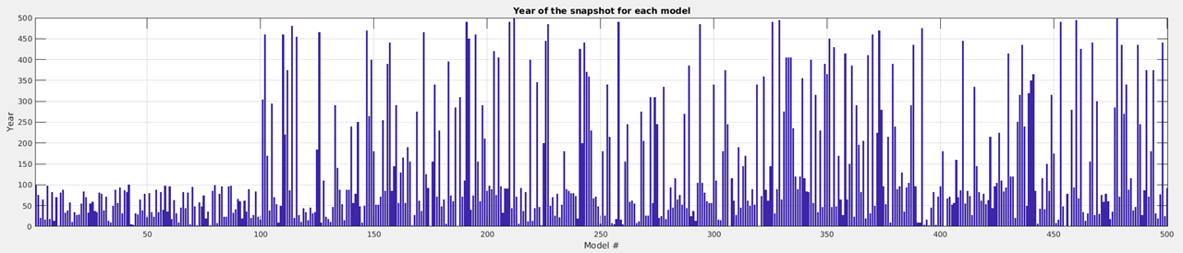}
    \caption{Distribution of time steps for our final training set of 500 CO$_2$ plume models. For the first 100 models, we randomly select time steps from the first 100 years of injection. For the remaining models, we randomly select time steps from all 500 years of simulation. \label{time-steps}}
\end{figure}

We select a single time step from each simulation to create our final dataset. For the first 100 simulations, we randomly select a time step from the first 100 years of injection. We randomly select time steps for the remaining 400 simulations across all 500 years. Figure \ref{time-steps} shows the distribution of the time steps in our final dataset. 

To reduce the computational cost (i.e., memory and time) and to accommodate a larger batch size during training for our DL-based methods, we resample each 3D plume volume from a size of 440$\times$530$\times$145 to a size of 128$\times$128$\times$128. 

% \begin{itemize}
%     \item How are these maps produced?
%     \item What variables are perturbed between each data point?
%     \item Citations: \cite{johansen1}
% \end{itemize}

% % Include plot for time distributions
% % Similarity matrix between gravity maps

% \subsection{Sn{\o}hvit Plume Volumes}
% \begin{itemize}
%     \item Citations:
% \end{itemize}

\section{Methods}
\label{sec:methods}
% \todo[inline]{since this section is long, we need a little intro and plan for the section}
% In this section, the traditional gravity modeling is introduced since it is used to generate ...
In this section, we first introduce traditional gravity modeling since it is used to generate the surface gravity maps that correspond to the CO$_2$ plumes described in Section \ref{sec:data}. We then describe conventional L2 inversion in Section \ref{sec:method-l2}. Section \ref{sec:method-ml-workflow}, outlines our proposed DL-based approach for surface gravity inversion. 

\subsection{Modeling Gravity} 
\label{sec:gravity}
Time-lapse surface gravity surveys are investigated to monitor the propagation of CO$_2$ plumes.

Given a density perturbation $\Delta \rho$, the gravity field observed at a station located at $\mathbf{r'}$ can be expressed as:
\begin{equation} \label{gravity_field}
	\mathbf{g}(\mathbf{r'})= \gamma \iiint_V \frac{\mathbf{r} - \mathbf{r'}}{| \mathbf{r} - \mathbf{r'} | } \Delta \rho (\mathbf{r}) dV
\end{equation}
where $\mathbf{r}$ is the spatial coordinate, $V$ is the volume of the reservoir, and $\gamma = 6.6738480 \times 10^{-11} \text{m}^3 \cdot \text{kg}^{-1} \cdot \text{s}^{-2}$ is Newton's gravitational constant.

In our application, we assume that only the vertical component of $\mathbf{g}$ is recorded at stations uniformly laid out along the seabed, every 500m in the $x$ and $y$-directions. 
To accurately account for the reservoir geometry, (\ref{gravity_field}) is discretized by means of a finite elements technique. This implementation is done under the GEOSX multi-physics framework \cite{geosx}.

%Given a 3D density difference map $\rho$ with dimensions $n_x$, $n_y$, and $n_z$ and with spatial resolution $h_x$, $h_y$, and $h_z$, we compute the change in the downward force of gravity $g_i$ at station $s_i$ located at $(x_i, y_i)$ as follows:
%\begin{align*}
%    g_i = \sum_{k = 1}^{n_x n_y n_z} G h_x h_y h_z \rho(x_k, y_k, z_k) \frac{z_k}{\sqrt{(x_i - x_k)^2 + (y_i - y_k)^2 + z_k^2}},
%\end{align*}

\begin{figure}[ht!]
    \centering
    \includegraphics[width = 1\textwidth]{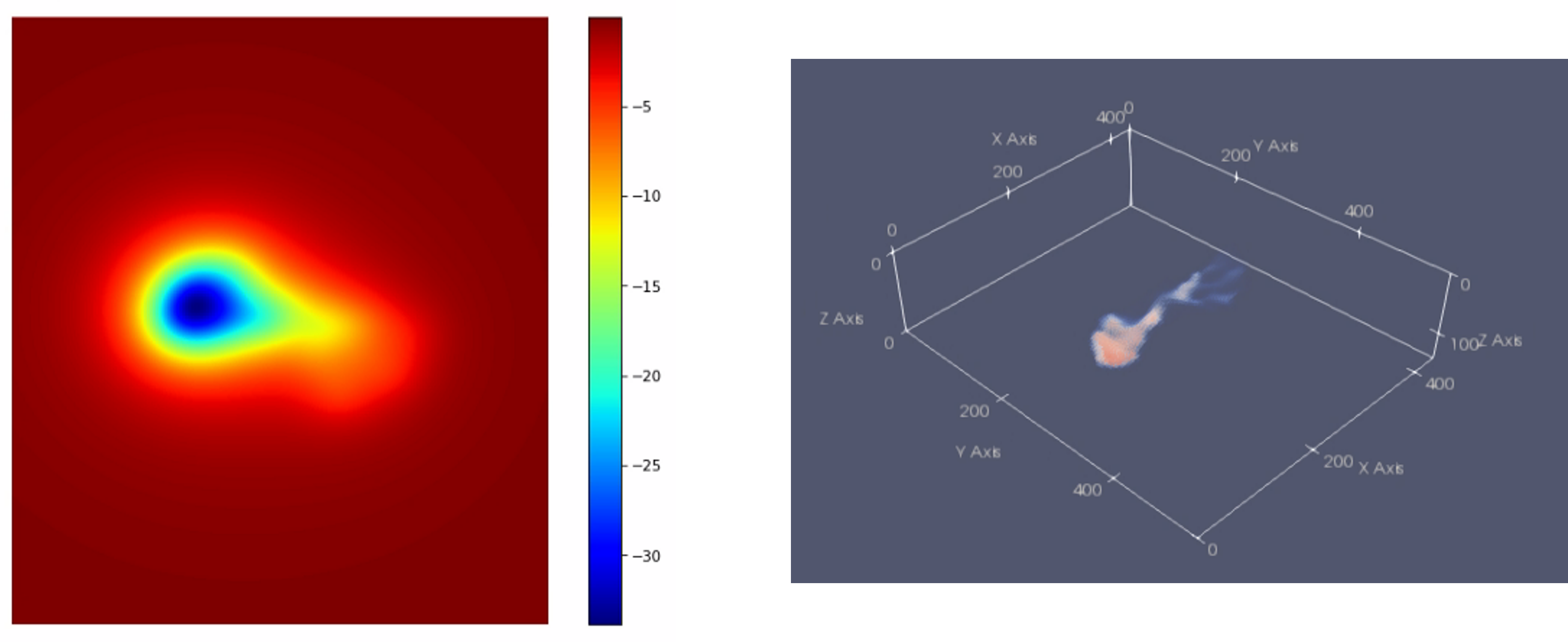}
    \caption{Computed 2D change in surface gravity map (left) for a given 3D difference in density map (right).\label{io-example}}
\end{figure}

%where $G = 6.6738480 \times 10^{-11} \text{m}^3 \cdot \text{kg}^{-1} \cdot \text{s}^{-2}$ is Newton's constant. We multiply the resultant $g_i$ by $1 \times 10^{-8}$ and report the result in $\mu$Gals. 
%We repeat this computation across a uniformly spaced grid where stations are placed every 500m to produce a 2D surface gravity map. 
Figure \ref{io-example} shows an example of a computed surface gravity difference map with its corresponding density difference volume. 
%We generate 500 density/gravity map pairs for training our deep learning model using the above methods.

Data normalization is an essential preprocessing step to achieve an acceptable predictive result for DL methods. We normalize each surface gravity map by subtracting its mean and dividing by its standard deviation (i.e., z-score normalization).

\subsection{L2 Inversion of Gravity Data}
\label{sec:method-l2}
The goal of many traditional inversion methods (i.e., non-DL-based methods) is to iteratively fit the gravity response from a predicted subsurface density anomaly to the original data. More specifically, the densities in a predicted model are found by minimizing an objective function subject to fitting the observed data. In the case of L2 inversion, we use an L2 formulation of the objective function. To define the objective function, we first let $F: \mathbb{R}^{n_1 \times n_2 \times n_3} \rightarrow \mathbb{R}^{m_1 \times m_2}$ be a forward gravity model that maps a density anomaly of size $n_1 \times n_2 \times n_3$ to a surface gravity response of size $m_1 \times m_2$. In its most basic form, the objective function is given by
\begin{align*}
    O(\rho) = \frac{1}{2} \norm{F(\rho) - G}^2,
\end{align*}
where $\rho$ is a subsurface density model, and $G$ is the observed data. Our goal then is to solve the following minimization problem
\begin{align} \label{eq:1}
    \min_{\rho} O(\rho).
\end{align}

\begin{figure}[ht!]
    %\centering
    \hspace{-.5in}
    \includegraphics[width = 1.2\textwidth]{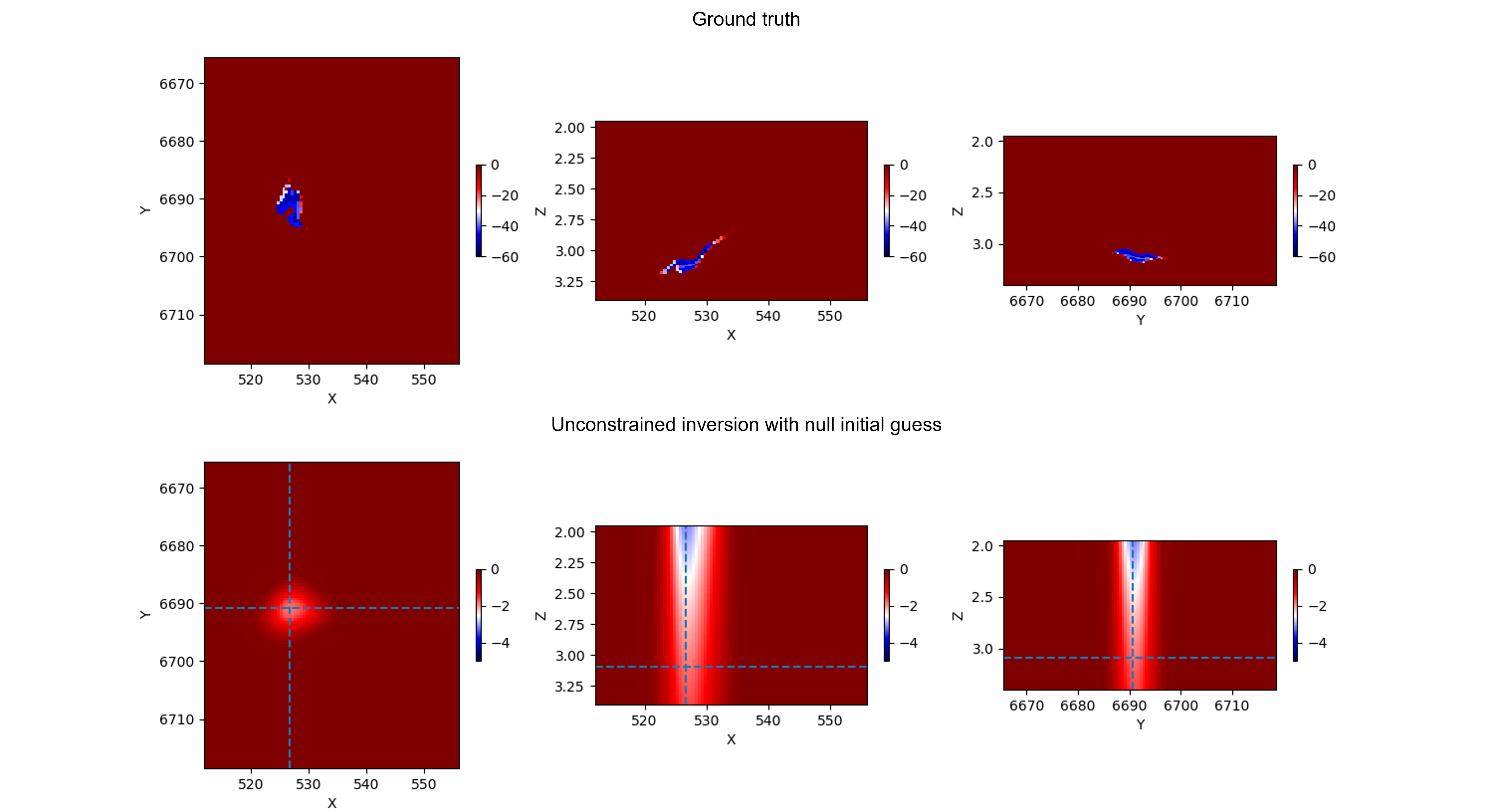}
    \caption{(Top) True CO$_2$ plume. (Bottom) Predicted CO$_2$ plume from unconstrained L2 inversion with a null model as an initial guess.\label{l2-unconstrained}}
\end{figure}

Due to the ill-posed nature of gravity inversion (i.e., several anomalies can produce the same gravity response), it is not sufficient to solve (\ref{eq:1}) in an unconstrained setting. Figure \ref{l2-unconstrained} illustrates the results of unconstrained L2 inversion with a null model as an initial guess. While the predicted density anomaly produces a gravity response that fits the original data, it does not match the true density anomaly. To overcome this challenge, we constrain (\ref{eq:1}) so that only cells within the known reservoir mask are updated during minimization. We use GEOSX for all L2 inversions \cite{geosx}.

\subsection{Inversion of Gravity Data with Deep Learning}
\label{sec:method-ml-workflow}
DL-based inversion is a data-driven method that builds a deep neural network model to map geophysical observations into a subsurface physical property model such as seismic velocity, electrical conductivity, density, or saturation. A large amount of data is required to train the deep neural network model. Once trained, the DL model can predict the physical property model from measured data in near real-time. In this study, a U-Net model was developed and used to predict subsurface CO$_2$ distribution from surface gravity monitoring data based on distributions from a physics simulation of the Johansen formation. This process consists of six steps:
\begin{enumerate}
    \item Create synthetic subsurface CO$_2$ plume models (Section \ref{sec:data}).
    \item Generate corresponding gravity data on the ground surface using a forward gravity model (Section \ref{sec:gravity}).
    \item Build a DL architecture (i.e., U-Net).
    \item Formulate a loss function to minimize during training.
    \item Select hyperparameters (i.e., batch size and learning rate) and train the U-Net with the surface gravity data as input and CO$_2$ plume volumes as output.
    \item Select metrics and assess the prediction accuracy of the U-Net model with test data. 
\end{enumerate}
% Stochastic training, pattern recognition

% Main elements are architecture, loss function, hyperparameters

\subsubsection{Deep Learning Architecture}
\label{sec:method-ml-model}
We propose a modified 3D U-Net \cite{unet, unet-3d} to invert 2D surface gravity maps into 3D plume density maps. At the start of our proposed architecture, we resize the 2D surface maps to match the height and width of the 3D plume volume via a series of 2D convolution layers and spline interpolation. We then convert the 2D surface maps into 3D volumes via a pointwise convolution, where the number of channels equals the depth dimension of the volumetric output. This resultant volume is the input to a standard 3D U-Net. The output of the U-Net splits into two separate branches, where each branch applies a series of convolution operations followed by batch normalization and a ReLU activation function. One branch is responsible for segmenting the CO$_2$ plume and the other for predicting the density values within the plume.

\begin{figure}[ht!]
    \centering
    \includegraphics[width = 1\textwidth]{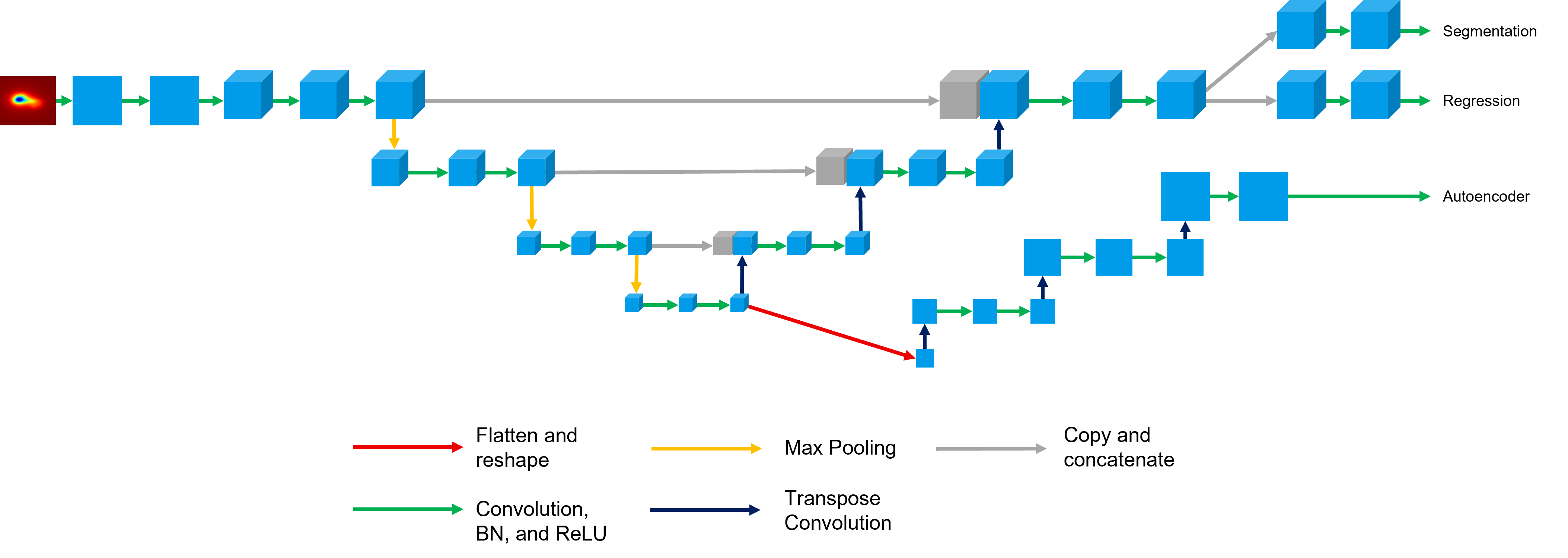}
    \caption{Sketch of our proposed architecture. \label{architecture}}
\end{figure}

Additionally, we apply autoencoder regularization to our network \cite{vae-reg}. The output of the U-Net's bottleneck is flattened, reshaped into a 2D array, and decoded to reproduce the input surface map. Figure \ref{architecture} provides a sketch of our deep learning architecture. Table \ref{tab:arch-details} provides implementation details for our proposed architecture.

% Talk about progression of architecture design
% Try with simpler architecture

\begin{table}[ht!]
\centering
\resizebox{\textwidth}{!}{%
\begin{tabular}{ll|ll|ll}
\hline
\multicolumn{2}{c|}{Input} &
  \multicolumn{2}{c|}{U-Net} &
  \multicolumn{2}{c}{Autoencoder} \\ \hline
\multicolumn{1}{c}{Block} &
  \multicolumn{1}{c|}{Description} &
  \multicolumn{1}{c}{Block} &
  \multicolumn{1}{c|}{Description} &
  \multicolumn{1}{c}{Block} &
  \multicolumn{1}{c}{Description} \\ \hline
Input &
  Input 2D surface gravity &
  Encoder &
  \begin{tabular}[c]{@{}l@{}}Repeat twice:\\ Conv3D(16, (3$\times$3$\times$3))\\ Batch Normalization\\ ReLU\\ \\ End with:\\ Maxpooling3D\end{tabular} &
  Reshape &
  \begin{tabular}[c]{@{}l@{}}Flatten output of Bottleneck \\ with GlobalMaxpooling3D\\ \\ Reshape to 4$\times$4$\times$16\\ \\ Conv2D(256, (3$\times$3))\end{tabular} \\ \hline
Resize &
  \begin{tabular}[c]{@{}l@{}}Resize gravity maps to \\ 128$\times$128 with\\ tf.keras.layers.Resizing \\ using lanczos5 interpolation\end{tabular} &
  \begin{tabular}[c]{@{}l@{}}Encoder\\ $\times$4\end{tabular} &
  \begin{tabular}[c]{@{}l@{}}Repeat encoder with \\ 32, 64, 128, and 256 filters\end{tabular} &
  Decoder &
  \begin{tabular}[c]{@{}l@{}}Conv2DTranspose(256, (3$\times$3))\\ \\ Repeat twice:\\ Conv2D(256, (3$\times$3))\\ Batch Normalization\\ ReLU\end{tabular} \\ \hline
Conv2DBlock &
  \begin{tabular}[c]{@{}l@{}}Conv2D(16, (7$\times$7))\\ Batch Normalization\\ ReLU\end{tabular} &
  Bottleneck &
  \begin{tabular}[c]{@{}l@{}}Repeat twice:\\ Conv3D(256, (3$\times$3$\times$3))\\ Batch Normalization\\ ReLU\end{tabular} &
  \begin{tabular}[c]{@{}l@{}}Decoder\\ $\times$4\end{tabular} &
  \begin{tabular}[c]{@{}l@{}}Repeat decoder with \\ 128, 64, 32, and 16 filters\end{tabular} \\ \hline
\begin{tabular}[c]{@{}l@{}}Conv2DBlock\\ $\times$2\end{tabular} &
  \begin{tabular}[c]{@{}l@{}}Conv2DBlock(16, (5$\times$5)) and\\ Con2DBlock(16, (3$\times$3))\end{tabular} &
  Decoder &
  \begin{tabular}[c]{@{}l@{}}Conv3DTranspose(256, (3$\times$3$\times$3))\\ \\ Concatenate with Encoder(256) output\\ \\ Repeat twice:\\ Conv3D(256, (3$\times$3$\times$3))\\ Batch Normalization\\ ReLU\end{tabular} &
  Output &
  Conv2D(1, (1$\times$1)) \\ \hline
Resize &
  \begin{tabular}[c]{@{}l@{}}Resize output of \\ last Conv2DBlock \\ to 128$\times$128 with \\ tf.keras.layers.Resizing \\ using lanczos5 interpolation\end{tabular} &
  \begin{tabular}[c]{@{}l@{}}Decoder\\ $\times$4\end{tabular} &
  \begin{tabular}[c]{@{}l@{}}Repeat dencoder with \\ 128, 64, 32, and 16 filters\end{tabular} &
   &
   \\ \hline
Add &
  Add first and last resize layers &
  Regression &
  \begin{tabular}[c]{@{}l@{}}Copy output of Decoder(16)\\ \\ Repeat twice:\\ Conv3D(16, (3$\times$3$\times$3))\\ Batch Normalization\\ ReLU\\ \\ Ends with:\\ Conv3D(1, (1$\times$1$\times$1))\end{tabular} &
   &
   \\ \hline
Conv2D &
  \begin{tabular}[c]{@{}l@{}}Convert 2D input to \\ 3D cube with\\ Conv2D(128, (1$\times$1))\end{tabular} &
  Segmentation &
  \begin{tabular}[c]{@{}l@{}}Copy output of Decoder(16)\\ \\ Repeat twice:\\ Conv3D(16, (3$\times$3$\times$3))\\ Batch Normalization\\ ReLU\\ \\ Ends with:\\ Conv3D(1, (1$\times$1$\times$1))\\ Sigmoid\end{tabular} &
   &
   \\ \hline
\end{tabular}%
}
\caption{Implementation details for each stage in our proposed U-Net architecture. We first resize the 2D input and convert it to a 3D volume via pointwise convolution. This volume is the input to a standard 3D U-Net described in the center column. The details of autoencoder regularization are shown in the right-most column. \label{tab:arch-details}}
\end{table}

\subsubsection{Loss Function}
\label{sec:method-ml-loss}
\paragraph{Segmentation Loss.} For the segmentation branch of our architecture, we use the Generalized Dice Loss (GDL) function proposed by Sudre et al. \cite{gdl}. This loss function improves upon the original Dice loss proposed by Milletari et al. \cite{vnet} by adding a weighting term for each segmentation class. Adding weight terms produced significantly better Dice scores for highly imbalanced segmentation problems. The GDL loss is given by the following:
\begin{align*}
    \mathcal{L}_{gdl} = 1 - 2\frac{\sum_{k=1}^{C} w_k \sum_{i=1}^{N} T_i^k P_i^k}{\sum_{k=1}^{C} w_k \sum_{i=1}^{N} (T_i^k)^2 + (P_i^k)^2},
    % \mathcal{L}_{gdl} = \frac{\sum_{k=1}^C w_k \norm{T^k - P^k}_2^2}{\sum_{k=1}^C w_k \left(\norm{T^k}_2^2 + \norm{P^k}_2^2\right)},
\end{align*}
where $C$ denotes the number of segmentation classes, $N$ denotes the total number of pixels (or voxels in the 3D case), $P_i^k$ is the $i^{th}$ voxel in the predicted segmentation mask for class $k$, and $T_i^k$ is the $i^{th}$ voxel in the ground truth mask for class $k$. The term $w_k$ is the weighting term for the $k^{th}$ class and is given by
\begin{align*}
    w_k = \left(\frac{C}{\sum_{k=1}^{C} \frac{1}{N_k}}\right) \frac{1}{N_k},
    % w_k = \frac{1}{\left( \sum_{i=1}^{N} T_i^k \right)^2 + 1}.
\end{align*}
where $N_k$ is the total number of pixels belonging to the class $k$ over the entire dataset. Note that the weights $w_k$ are pre-computed and remain constant throughout training. 

In our case, the number of segmentation classes equals two; background and foreground. The computed class weights are approximately 0.003 and 1.997 for the background and foreground classes, respectively.

\paragraph{Regression Loss.} For the regression branch of our architecture, we use the mean squared error loss function, which is given by
\begin{align*}
    \mathcal{L}_{reg} = \frac{1}{N}\norm{\rho_T - \rho_P}_2^2,
\end{align*}
where $\rho_T$ and $\rho_P$ are the ground truth and predicted density maps, respectively.

\paragraph{Autoencoder Loss.} For the autoencoder branch of our architecture, we, again, use the mean squared error loss function. This loss is given by
\begin{align*}
    \mathcal{L}_{ae} = \frac{1}{N}\norm{g_T - g_P}_2^2,
\end{align*}
where $g_T$ and $g_P$ are the ground truth and reconstructed surface gravity maps, respectively.

\paragraph{Composite Loss.} We take a weighted combination of the segmentation, regression, and autoencoder loss functions to produces our final loss for training. This loss function is given by
\begin{align*}
    \mathcal{L} = 0.7\mathcal{L}_{reg} + 0.25\mathcal{L}_{gdl} + 0.05\mathcal{L}_{ae}.
\end{align*}

\subsubsection{Training and Testing Protocols}
\label{sec:method-ml-params}
We initialize the first layer in our modified U-Net architecture with 16 feature maps and use the Adam optimizer \cite{adam}. The initial learning rate is set at 0.001 with a cosine decay schedule with restarts \cite{cosine-lr}. During training, we use a batch size of eight. Random flips and additive Gaussian noise are applied to each batch as data augmentation. To evaluate the validity of our predicted inversions, we utilize the following metrics:

\begin{itemize}
    \item Mean Squared Error (kg/m$^3$) - The mean squared error in kg/m$^3$ between the predicted and true density anomalies.
    
    \item Mean Squared Error ($\mu$Gal) - The mean squared error in $\mu$Gal between the input gravity map and the gravity response from the predicted anomaly.
    
    \item R-Squared - The R-squared coefficient between the predicted and true density anomalies. 
    
    \item Dice Similarity Coefficient - A measure of overlap between the non-zero masks of the predicted and true density anomalies. The Dice similarity coefficient (Dice) ranges from zero to one, where one represents a perfect prediction \cite{dice}.
\end{itemize}

Unless otherwise specified, all experiments utilize the same 90/10 train-test split with 5\% of the training set used as a validation set. This split results in a training set with 450 gravity/plume pairs and a test set with 50 pairs. Our models are implemented in Python using TensorFlow (v2.8.0) and trained on four NVIDIA A100 GPUs \cite{keras}. We leave all other hyperparameters at their default values.

At test time, our DL-based inversion produces predictions that are of size 128$\times$128$\times$128. To produce our final prediction, we resample this output to the original grid resolution of 440$\times$530$\times$145 via linear interpolation. 

\section{Results}
\label{sec:eval}

\subsection{L2 Inversion}
\begin{figure}[ht!]
    \centering
    \includegraphics[width = 1\textwidth]{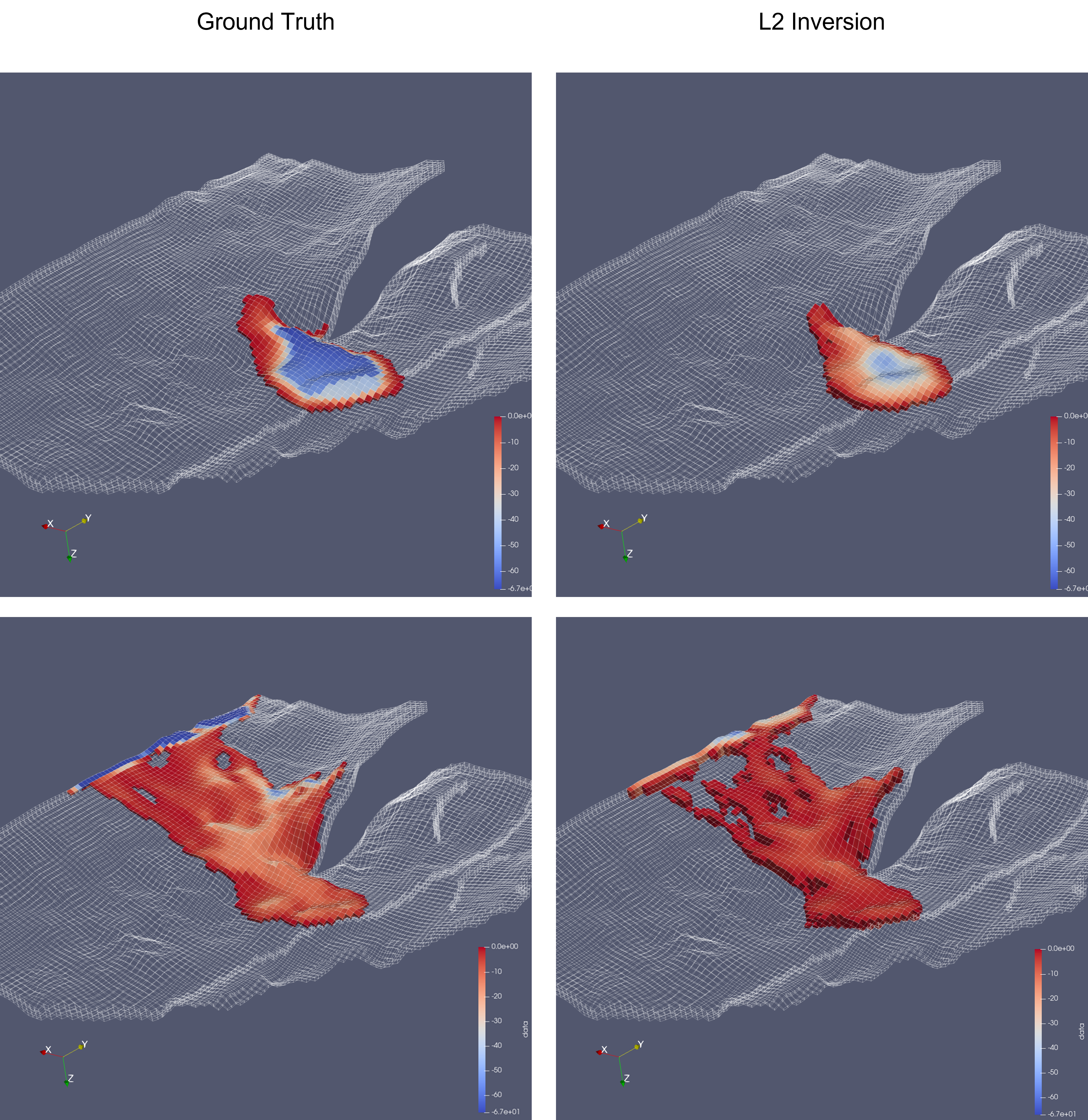}
    \caption{(Left) True CO$_2$ plume. (Right) Predicted CO$_2$ plume using contrained L2 inversion.\label{l2-preds}}
\end{figure}

Using the methods described in Section \ref{sec:method-l2}, we invert the 50 surface gravity maps in the test set described in \ref{sec:method-ml-params}. Table \ref{tab:results-l2-table} reports the accuracy of these inversions using the metrics described in Section \ref{sec:method-ml-params}. Figure \ref{l2-preds} illustrates two predicted CO$_2$ plumes using L2 inversion.

\setlength\tabcolsep{10pt}
\begin{table}[ht!]
\centering
\bgroup
\def\arraystretch{1.25}%  1 is the default, change whatever you need
\begin{tabular}{lcccc}
\hline
\multirow{2}{*}{} & \multicolumn{1}{l}{\multirow{2}{*}{Mean $\pm$ Std}} & \multicolumn{1}{l}{\multirow{2}{*}{Median}} & \multicolumn{2}{c}{Percentile} \\ \cline{4-5} 
                & \multicolumn{1}{l}{} & \multicolumn{1}{l}{} & 25$^\text{th}$ & 75$^\text{th}$ \\ \hline
MSE (kg/m$^3$)  & 0.67 $\pm$ 0.39      & 0.72                 & 0.35           & 0.93          \\ \hline
MSE ($\mu$Gal)  & 0.00 $\pm$ 0.00      & 0.00                 & 0.00           & 0.00          \\ \hline
R-Squared       & -0.25 $\pm$ 0.73     & -0.01                & -0.44          & 0.20          \\ \hline
Dice            & 0.44 $\pm$ 0.06      & 0.45                 & 0.39           & 0.48          \\ \hline
\end{tabular}
\egroup
\caption{Accuracy of L2 inversion of surface gravity data.}
\label{tab:results-l2-table}
\end{table}

\subsection{Deep Learning Inversion}
\subsubsection{Data-Driven Deep Learning Inversion}
\begin{figure}[ht!]
    \centering
    \includegraphics[width = 1\textwidth]{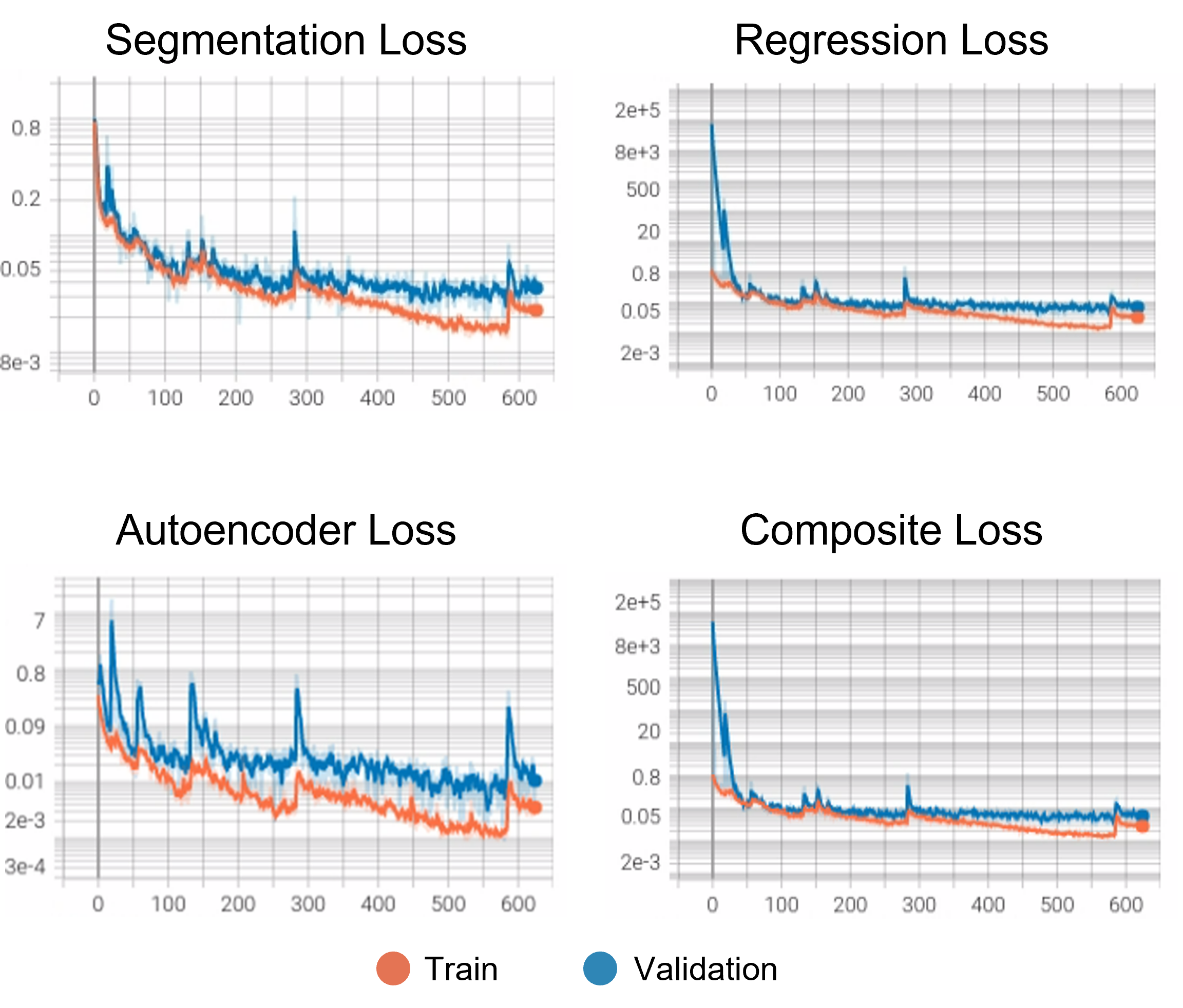}
    \caption{(Top left) Learning curves for segmentation loss. (Top right) Learning curves for regression loss. (Bottom left) Learning curves for autoencoder loss. (Bottom right) Learning curves for composite loss. All plots are plotted using a logarithmic scale. In each case, we see convergence in both the training and validation curves, which indicates that our DL approach successfully learned a mapping from surface gravity to 3D subsurface CO$_2$ plumes. \label{losses}}
\end{figure}

Using the methods described in Sections \ref{sec:method-ml-workflow} - \ref{sec:method-ml-params}, we train our proposed DL architecture on 450 gravity/plume pairs. We randomly set aside 5\% of the training pairs as a validation set. The loss curves for each constituent loss and the composite loss described in \ref{sec:method-ml-loss} are shown in Figure \ref{losses}. We see convergence for the training and validation sets for each loss function. This convergence indicates that our modified U-Net was able to successfully learn a mapping from observed surface gravity data to 3D subsurface CO$_2$ plumes.

\begin{figure}[ht!]
    \centering
    \includegraphics[width = 1\textwidth]{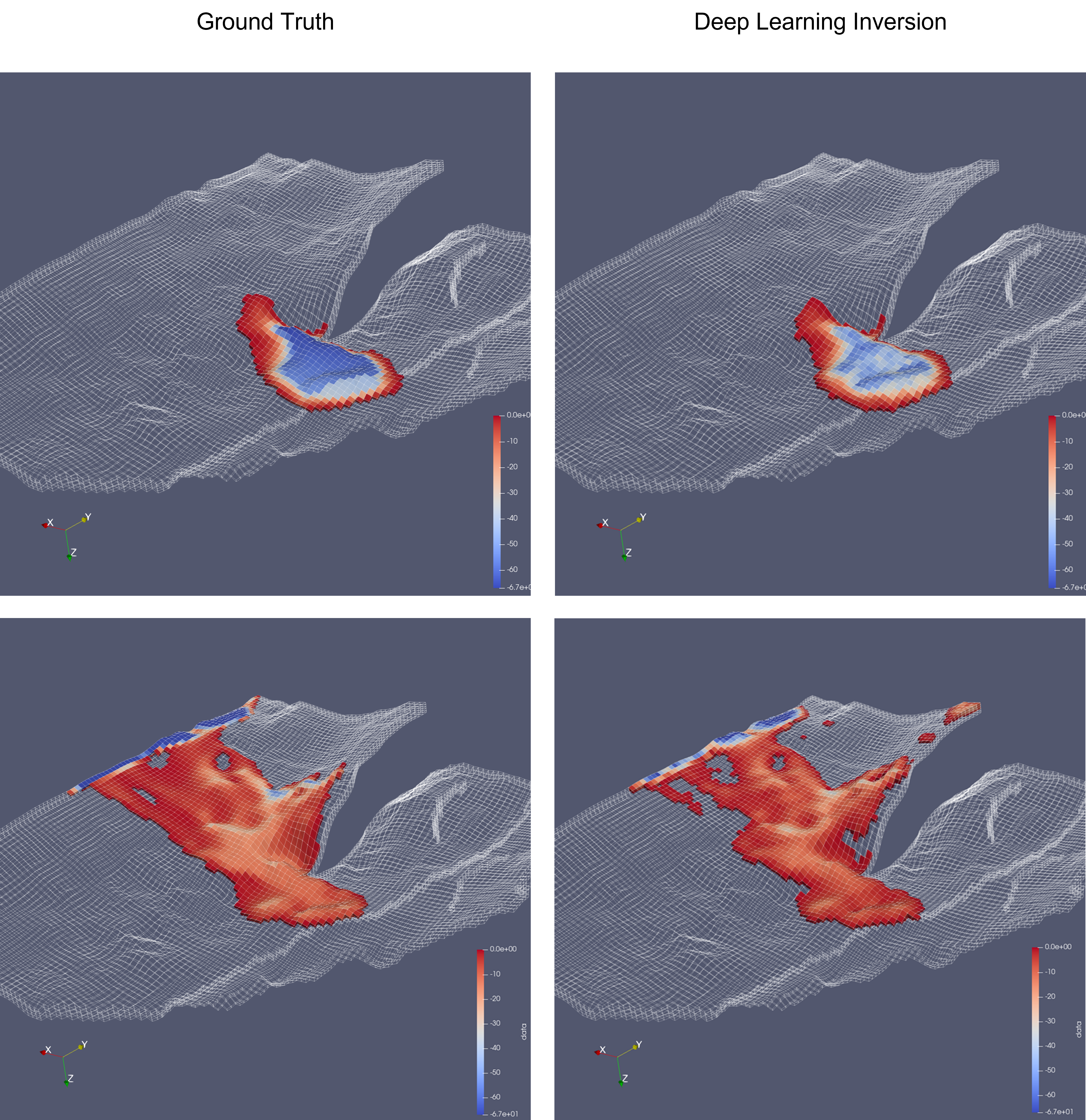}
    \caption{(Left) True CO$_2$ plume. (Right) Predicted CO$_2$ plume using deep learning.\label{dl-preds}}
\end{figure}

To assess the accuracy of our trained DL model, we, again, invert the 50 surface gravity maps in our fixed test set and compute the metrics described in Section \ref{sec:method-ml-params} for the predicted CO$_2$ plumes. These metrics are shown in Table \ref{tab:results-dl}. In Table \ref{tab:results-dl} we see that our DL approach outperforms L2 inversion in terms of plume misfit (i.e., MSE (kg/m$^3$) and R-Squared) and plume geometry (i.e., Dice), but does not outperform L2 inversion in terms of data misfit (i.e., MSE ($\mu$Gal)). This latter underperformance can be explained by the fact that our DL model does not attempt to fit the gravity response of the predicted plume to the original data. Instead, our proposed DL approach directly fits the predicted CO$_2$ plume to the true CO$_2$ plume. Figure \ref{dl-preds} illustrates two predicted CO$_2$ plumes using DL-based inversion.

\setlength\tabcolsep{10pt}
\begin{table}[ht!]
\centering
\bgroup
\def\arraystretch{1.25}%  1 is the default, change whatever you need
\begin{tabular}{lcccc}
\hline
\multirow{2}{*}{} & \multicolumn{1}{l}{\multirow{2}{*}{Mean $\pm$ Std}} & \multicolumn{1}{l}{\multirow{2}{*}{Median}} & \multicolumn{2}{c}{Percentile} \\ \cline{4-5} 
                & \multicolumn{1}{l}{} & \multicolumn{1}{l}{} & 25$^\text{th}$ & 75$^\text{th}$ \\ \hline
MSE (kg/m$^3$)  & 0.26 $\pm$ 0.15      & 0.27                 & 0.14           & 0.36          \\ \hline
MSE ($\mu$Gal) & 0.47 $\pm$ 0.89      & 0.13                 & 0.03           & 0.50          \\ \hline
R-Squared           & 0.68 $\pm$ 0.17      & 0.73                 & 0.62           & 0.80          \\ \hline
Dice            & 0.78 $\pm$ 0.03      & 0.79                 & 0.76           & 0.80          \\ \hline
\end{tabular}
\egroup
\caption{Accuracy of deep learning-based inversion of surface gravity data. Here we see that our proposed DL method outperforms L2 inversion in terms of CO$_2$ plume error but does not match the gravity response of the predicted plume with the observed data.}
\label{tab:results-dl}
\end{table}

% \begin{figure}[ht!]
%     \centering
%     \includegraphics[width = 1\textwidth]{images/boxplots.png}
%     \caption{(Top Left) Distribution of mean squared error for predicted CO$_2$ plumes in kg/m$^3$. (Top Right) Distribution of R-squared coefficient for predicted CO$_2$ plumes. (Bottom Left) Distribution of mean squared error for gravity response of predicted CO$_2$ plumes in $\mu$Gals. (Bottom Right) Distribution of Dice coefficients for non-zero mask of predicted CO$_2$ plumes. \label{boxplots}}
% \end{figure}

We also performed a five-fold cross-validation to assess the robustness of our proposed DL-based inversion. In this scheme, we use 20\% of the dataset for testing and train on the remaining 80\% of the data. We, again, set aside 5\% of the training data as a validation set. This process iterates across the entire dataset until we have test time predictions for all 500 surface gravity/plume pairs. Table \ref{tab:results-dl-five-fold} shows the results of the five-fold cross-validation. We see that these results are nearly identical to the results shown in Table \ref{tab:results-dl}, which indicates that our DL model learned a similar mapping from surface gravity to subsurface CO$_2$ plumes across multiple perturbations of the training data. 

\setlength\tabcolsep{10pt}
\begin{table}[ht!]
\centering
\bgroup
\def\arraystretch{1.25}%  1 is the default, change whatever you need
\begin{tabular}{lcccc}
\hline
\multirow{2}{*}{} & \multicolumn{1}{l}{\multirow{2}{*}{Mean $\pm$ Std}} & \multicolumn{1}{l}{\multirow{2}{*}{Median}} & \multicolumn{2}{c}{Percentile} \\ \cline{4-5} 
                & \multicolumn{1}{l}{} & \multicolumn{1}{l}{} & 25$^\text{th}$ & 75$^\text{th}$ \\ \hline
MSE (kg/m$^3$)  & 0.27 $\pm$ 0.16      & 0.26                 & 0.14           & 0.35          \\ \hline
MSE ($\mu$Gal) & 0.43 $\pm$ 1.12      & 0.07                 & 0.01           & 0.24          \\ \hline
R-Squared           & 0.69 $\pm$ 0.22      & 0.74                 & 0.66           & 0.80          \\ \hline
Dice            & 0.75 $\pm$ 0.05      & 0.76                 & 0.73           & 0.78          \\ \hline
\end{tabular}
\egroup
\caption{Accuracy of deep learning-based inversion of surface gravity data using a five-fold cross validation.}
\label{tab:results-dl-five-fold}
\end{table}

\subsubsection{Combined L2-Deep Learning Inversion}
To overcome the shortfalls of L2 inversion (i.e., large model misfit) and DL-based inversion (i.e., large data misfit), we use L2 inversion as a postprocessing step for our DL-driven inversion. After generating a prediction with our DL model, we use that prediction as an initial guess for L2 inversion. Table \ref{tab:results-l2dl} shows the results of this process. Our combined approach produces predictions whose gravity response matches the original observation and further reduces the model misfit. The only metric that does not improve is the Dice coefficient. This decrease in performance for Dice can be explained by the fact that our constrained approach to L2 inversion tends to ``fill'' a large number of grid cells within the reservoir mask with small non-zero values to fit the gravity response with the original data. This process produces non-zero masks much larger than the true non-zero mask for the CO$_2$ plume, which degrades the Dice metric. As a postprocessing step, we can apply a  threshold to the combined L2-DL prediction to improve the Dice metric. We found via a grid search that the optimal threshold is -7 kg/m$^3$. This threshold improves the mean Dice coefficient from 0.41 to 0.62.

\begin{figure}[ht!]
    \centering
    \includegraphics[width = 1\textwidth]{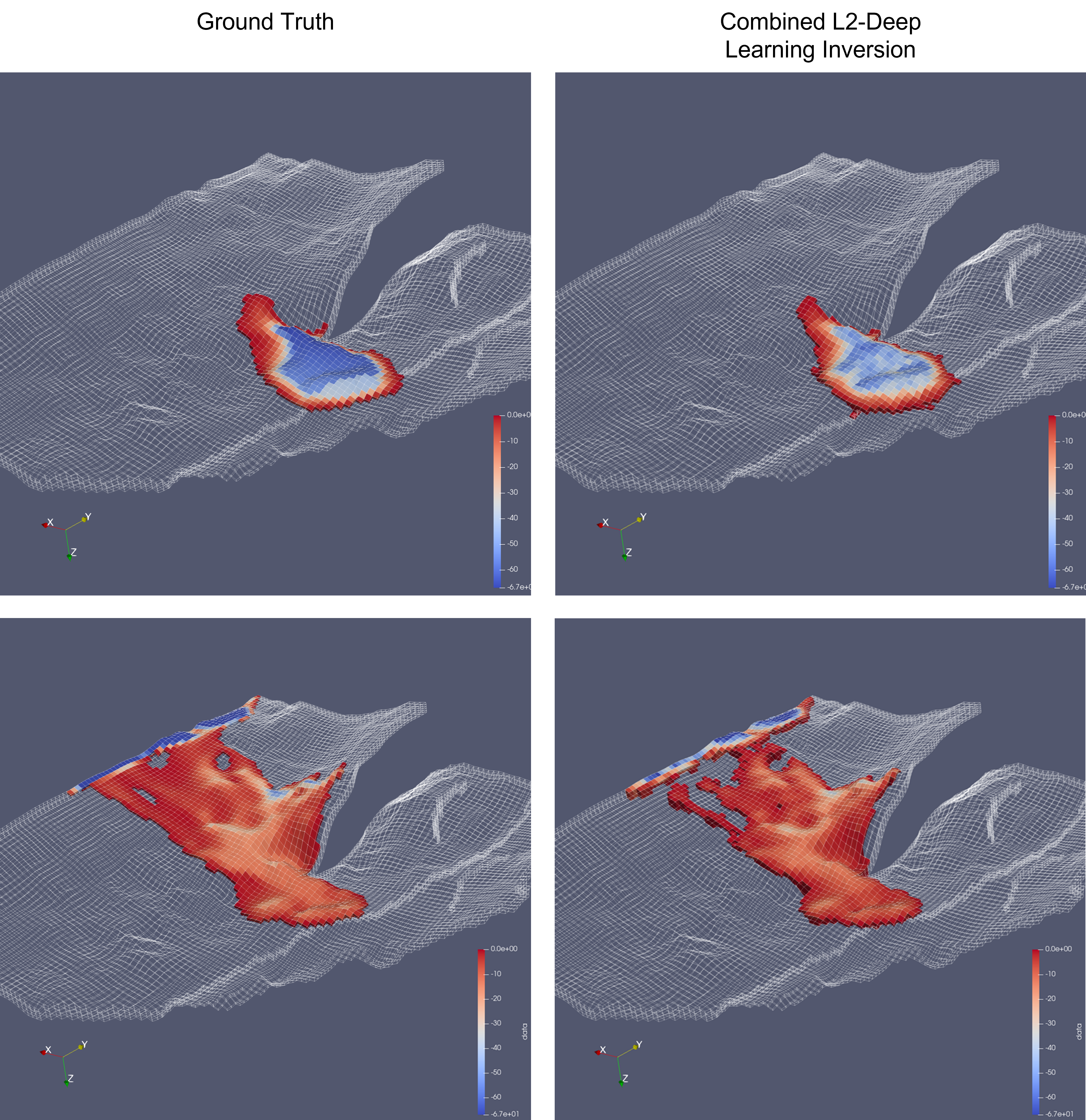}
    \caption{(Left) True CO$_2$ plume. (Right) Predicted CO$_2$ plume using deep learning prediction as an initial guess for L2 inversion.\label{l2dl_preds}}
\end{figure}

Figure \ref{l2dl_preds} illustrates two predicted CO$_2$ plumes using DL-based inversion. Visually, there is little difference between the smaller plume shown at the top of the figure and the original DL prediction, but our combined L2-DL method appears to improve the prediction performance for larger CO$_2$ plumes.

\setlength\tabcolsep{10pt}
\begin{table}[ht!]
\centering
\bgroup
\def\arraystretch{1.25}%  1 is the default, change whatever you need
\begin{tabular}{lcccc}
\hline
\multirow{2}{*}{} & \multicolumn{1}{l}{\multirow{2}{*}{Mean $\pm$ Std}} & \multicolumn{1}{l}{\multirow{2}{*}{Median}} & \multicolumn{2}{c}{Percentile} \\ \cline{4-5} 
                & \multicolumn{1}{l}{} & \multicolumn{1}{l}{} & 25$^\text{th}$ & 75$^\text{th}$ \\ \hline
MSE (kg/m$^3$)  & 0.13 $\pm$ 0.09      & 0.11                 & 0.07           & 0.17          \\ \hline
MSE ($\mu$Gal)  & 0.00 $\pm$ 0.00      & 0.00                 & 0.00           & 0.00          \\ \hline
R-Squared       & 0.89 $\pm$ 0.06      & 0.90                 & 0.86           & 0.94          \\ \hline
Dice            & 0.41 $\pm$ 0.09      & 0.43                 & 0.38           & 0.47          \\ \hline
\end{tabular}
\egroup
\caption{Accuracy of combined L2-Deep Learning approach for gravity inversion. Our combined approach produces predictions whose gravity response matches the original observation and further reduces the model misfit. \label{tab:results-l2dl}}
\end{table}

\subsubsection{Deep Learning Inversion With Physics-Driven Workflow}
\begin{figure}[ht!]
    \centering
    \includegraphics[width = 1\textwidth]{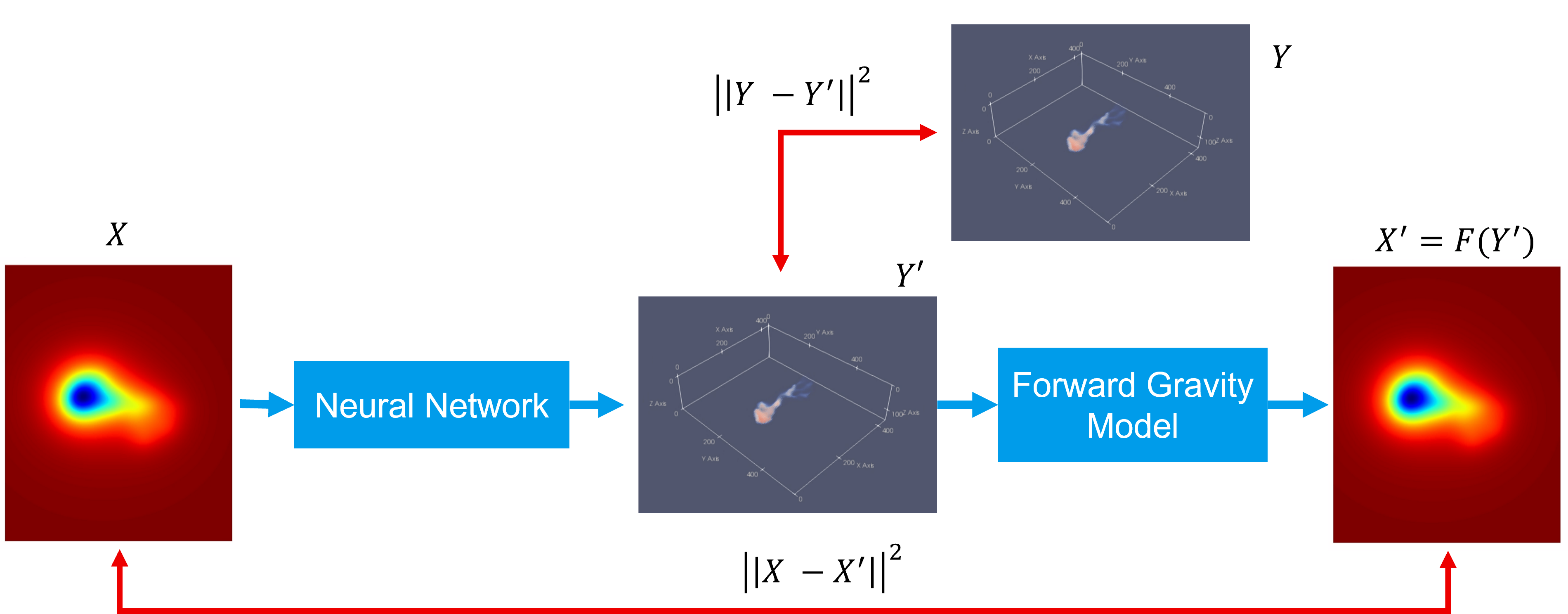}
    \caption{A sketch of a proposed physics-informed approach to training DL models for surface gravity inversion. Here, the neural network iteratively minimizes the model and data misfit. \label{workflow}}
\end{figure}

Another method we propose to mitigate the data misfit seen in our pure DL approach is a hybrid DL-aided geophysical inversion approach that iteratively minimizes model and data misfit during training. Given the simplicity and small computational cost associated with gravity modeling, we can append our forward gravity model to our neural network and create a hybrid workflow to minimize model and data misfit. Figure \ref{workflow} shows a sketch of our proposed workflow. 

Table \ref{tab:results-hybrid} shows the results of training a model using the workflow described above. In general, we see that these results match those of our purely DL-driven approach. The only metric that does improve is the R-Squared coefficient. 

\setlength\tabcolsep{10pt}
\begin{table}[ht!]
\centering
\bgroup
\def\arraystretch{1.25}%  1 is the default, change whatever you need
\begin{tabular}{lcccc}
\hline
\multirow{2}{*}{} & \multicolumn{1}{l}{\multirow{2}{*}{Mean $\pm$ Std}} & \multicolumn{1}{l}{\multirow{2}{*}{Median}} & \multicolumn{2}{c}{Percentile} \\ \cline{4-5} 
                & \multicolumn{1}{l}{} & \multicolumn{1}{l}{} & 25$^\text{th}$ & 75$^\text{th}$ \\ \hline
MSE (kg/m$^3$)  & 0.26 $\pm$ 0.15      & 0.26                 & 0.15           & 0.33          \\ \hline
MSE ($\mu$Gal)  & 0.64 $\pm$ 1.19      & 0.24                 & 0.05           & 0.77          \\ \hline
R-Squared       & 0.79 $\pm$ 0.07      & 0.81                 & 0.78           & 0.83          \\ \hline
Dice            & 0.77 $\pm$ 0.06      & 0.78                 & 0.76           & 0.80          \\ \hline
\end{tabular}
\egroup
\caption{Accuracy of hybrid training approach for gravity inversion. We generally see similar results to our pure DL-driven approach to inversion. The only metric that benefits is the R-Squared coefficient. \label{tab:results-hybrid}}
\end{table}

Additionally, we use transfer learning with our hybrid approach. More specifically, we take our converged, purely data-driven DL-driven model and insert it into our physics-driven workflow. This modification significantly speeds up the time required for our workflow to converge. Without transfer learning, our workflow takes roughly four days to converge. With transfer learning, we see convergence in eight hours. Table \ref{tab:results-hybrid-transfer} shows the transfer learning results with our workflow. Here we see a slight improvement in our data misfit and similar metrics to our non-transfer learning-based approach and a comparable improvement in R-Squared versus DL-based inversion. 

\setlength\tabcolsep{10pt}
\begin{table}[ht!]
\centering
\bgroup
\def\arraystretch{1.25}%  1 is the default, change whatever you need
\begin{tabular}{lcccc}
\hline
\multirow{2}{*}{} & \multicolumn{1}{l}{\multirow{2}{*}{Mean $\pm$ Std}} & \multicolumn{1}{l}{\multirow{2}{*}{Median}} & \multicolumn{2}{c}{Percentile} \\ \cline{4-5} 
                & \multicolumn{1}{l}{} & \multicolumn{1}{l}{} & 25$^\text{th}$ & 75$^\text{th}$ \\ \hline
MSE (kg/m$^3$)  & 0.25 $\pm$ 0.15      & 0.25                 & 0.14           & 0.32          \\ \hline
MSE ($\mu$Gal)  & 0.43 $\pm$ 0.85      & 0.13                 & 0.02           & 0.45          \\ \hline
R-Squared       & 0.80 $\pm$ 0.08      & 0.82                 & 0.79           & 0.84          \\ \hline
Dice            & 0.73 $\pm$ 0.07      & 0.75                 & 0.72           & 0.76         \\ \hline
\end{tabular}
\egroup
\caption{Accuracy of hybrid training approach for gravity inversion using transfer learning with converged pure DL-driven model. We see that our data misfit slightly improves (i.e., MSE ($\mu$Gal)) compared to training our physics workflow from scratch and our pure DL-based approach. \label{tab:results-hybrid-transfer}}
\end{table}

\subsubsection{Deep Learning Inversion With Time-Dependence}
All the methods tested so far use a single gravity map to produce a predicted reconstruction of the underlying subsurface density model. These methods, however, do not consider an underlying time dependence associated with our surface gravity data. We propose adding a convolutional LSTM network at the start of our DL architecture to process sequences of gravity data. Table \ref{tab:lstm-arch} gives the implementation details of this LSTM network. To train this time-dependent network, we take sequences of length ten from a simulation of the Johansen formation. More specifically, for the gravity map at time $t_9$, we prepend the data for time steps $t_0, \dots, t_8$ to make a single training example. To create our training dataset, we repeat this process for each subsequent time step in our simulation (i.e., $t_{10}, \dots$). We train our time-dependent network using the same protocols described in Section \ref{sec:method-ml-workflow}. Table \ref{tab:results-time} shows the results of training a time-dependent model. Here we see that our time-dependent DL approach can successfully recover our test cases with high accuracy in terms of model and data misfit. 

% \usepackage{graphicx}
\begin{table}[]
\centering
\resizebox{\textwidth}{!}{%
\begin{tabular}{lll}
\hline
Layer & Description                             & Output Size                                \\ \hline
Input & Sequences of gravity data of length ten & 1$\times$10$\times$128$\times$128$\times$1 \\ \hline
ConvLSTM2D &
  \begin{tabular}[c]{@{}l@{}}Use tf.keras.layers.ConvLSTM2D with \\ 16 features maps and kernel size seven.\\ \\ Follow with ReLU activation and batch\\ normalization\end{tabular} &
  1$\times$10$\times$128$\times$128$\times$16 \\ \hline
\begin{tabular}[c]{@{}l@{}}ConvLSTM2D\\ $\times$2\end{tabular} &
  \begin{tabular}[c]{@{}l@{}}Repeat ConvLSTM2D with kernel sizes\\ five and three.\end{tabular} &
  1$\times$10$\times$128$\times$128$\times$16 \\ \hline
Conv3D &
  \begin{tabular}[c]{@{}l@{}}Use tf.keras.layers.Conv3D with 128 \\ feature maps, kernel size one, and \\ channels first format to convert LSTM\\ output to volumetric cube for U-Net.\end{tabular} &
  1$\times$128$\times$128$\times$128$\times$1 \\ \hline
\begin{tabular}[c]{@{}l@{}}DL-Based \\ Inversion\\ U-Net\end{tabular} &
  \begin{tabular}[c]{@{}l@{}}See U-Net column in Table \ref{tab:arch-details} \\ \\ Output segmentation \\ and regression maps only\end{tabular} &
  \begin{tabular}[c]{@{}l@{}}1$\times$128$\times$128$\times$128$\times$1\\ $\times$2\end{tabular} \\ \hline
\end{tabular}%
}
\caption{Implementation details for our time-dependent DL-based inversion network. Here, we use an LSTM network to handle sequences of gravity maps before using the U-Net described in Section \ref{sec:method-ml-workflow}. \label{tab:lstm-arch}}
\end{table}

\setlength\tabcolsep{10pt}
\begin{table}[ht!]
\centering
\bgroup
\def\arraystretch{1.25}%  1 is the default, change whatever you need
\begin{tabular}{lcccc}
\hline
\multirow{2}{*}{} & \multicolumn{1}{l}{\multirow{2}{*}{Mean $\pm$ Std}} & \multicolumn{1}{l}{\multirow{2}{*}{Median}} & \multicolumn{2}{c}{Percentile} \\ \cline{4-5} 
                & \multicolumn{1}{l}{} & \multicolumn{1}{l}{} & 25$^\text{th}$ & 75$^\text{th}$ \\ \hline
MSE (kg/m$^3$)  & 0.10 $\pm$ 0.04      & 0.12                 & 0.06           & 0.13          \\ \hline
MSE ($\mu$Gal)  & 0.04 $\pm$ 0.03      & 0.03                 & 0.01           & 0.05          \\ \hline
R-Squared       & 0.92 $\pm$ 0.01      & 0.93                 & 0.92           & 0.93          \\ \hline
Dice            & 0.94 $\pm$ 0.01      & 0.95                 & 0.94           & 0.95          \\ \hline
\end{tabular}
\egroup
\caption{Results of DL-based inversion with time-dependence. Here we see that our time-dependent DL approach can successfully recover our test cases with high accuracy in terms of model and data misfit. \label{tab:results-time}}
\end{table}
% \begin{figure}[ht!]
%     \centering
%     \includegraphics[width = 1\textwidth]{images/example_preds.png}
%     \caption{From left to right - surface gravity maps, ground truth, segmentation prediction, and regression prediction for selected test set cases. \label{preds}}
% \end{figure}

\subsubsection{Deep Learning Inversion with Varying Sensor Resolution}
Up to this point, our results use gravity maps with a uniform sensor grid spacing of 500m. To further assess the robustness of our DL-based inversion, we generate surface gravity with sensors placed every 100m, 250m, 500m, 1km, 2km, and 3km. We train our DL-based inversion for each sensor grid resolution and report the results in Table \ref{tab:results-res}. The similarity of the results for each resolution indicates that our DL-based inversion can learn a mapping from surface gravity to subsurface changes in density for coarse and fine sensor grids.

\begin{table}[ht!]
\centering
\bgroup
\def\arraystretch{1.25}%  1 is the default, change whatever you need
\resizebox{\textwidth}{!}{%
\begin{tabular}{ccc|cc|cc|cc}
\hline
\multirow{2}{*}{\begin{tabular}[c]{@{}c@{}}Sensor\\ Resolution\end{tabular}} &
  \multicolumn{2}{c|}{MSE (kg/m$^3$)} &
  \multicolumn{2}{c|}{MSE ($\mu$Gal)} &
  \multicolumn{2}{c|}{R-Squared} &
  \multicolumn{2}{c}{Dice} \\ \cline{2-9} 
      & Mean $\pm$ Std  & Median & Mean $\pm$ Std  & Median & Mean $\pm$ Std  & Median & Mean $\pm$ Std  & Median \\ \hline
100m  & 0.26 $\pm$ 0.16 & 0.27   & 0.51 $\pm$ 1.03 & 0.11   & 0.66 $\pm$ 0.22 & 0.72   & 0.77 $\pm$ 0.04 & 0.78   \\
250m  & 0.26 $\pm$ 0.16 & 0.26   & 0.53 $\pm$ 1.01 & 0.11   & 0.67 $\pm$ 0.18 & 0.74   & 0.78 $\pm$ 0.03 & 0.79   \\
500m  & 0.26 $\pm$ 0.15 & 0.27   & 0.47 $\pm$ 0.89 & 0.13   & 0.68 $\pm$ 0.17 & 0.73   & 0.78 $\pm$ 0.03 & 0.79   \\
1000m & 0.27 $\pm$ 0.16 & 0.27   & 0.54 $\pm$ 1.12 & 0.11   & 0.66 $\pm$ 0.19 & 0.71   & 0.77 $\pm$ 0.05 & 0.78   \\
2000m & 0.27 $\pm$ 0.17 & 0.28   & 0.65 $\pm$ 1.23 & 0.22   & 0.63 $\pm$ 0.28 & 0.72   & 0.76 $\pm$ 0.06 & 0.78   \\
3000m & 0.27 $\pm$ 0.17 & 0.26   & 0.65 $\pm$ 1.22 & 0.22   & 0.66 $\pm$ 0.22 & 0.72   & 0.77 $\pm$ 0.04 & 0.78   \\ \hline
\end{tabular}%
}
\egroup
\caption{Accuracy of deep learning-based inversion for varying sensor grid resolutions. The similarity of the results for each resolution indicates that our DL-based inversion can learn a mapping from surface gravity to subsurface changes in density for coarse and fine sensor grids.}
\label{tab:results-res}
\end{table}

% \begin{figure}[ht!]
%     \centering
%     \includegraphics[width = 1\textwidth]{images/resolution.png}
%     \caption{Predictions for various sensor grid resolutions. \label{res}}
% \end{figure}

% \begin{figure}[ht!]
%     \centering
%     \includegraphics[width = 1\textwidth]{images/comp.png}
%     \caption{(Left) Ground truth, (Center) Deep learning prediction, and (Left) L2 inversion prediction. \label{methods}}
% \end{figure}

\subsubsection{Deep Learning Inversion with Saturation as Target}

All of our results so far demonstrate the ability of DL-based inversion to map surface gravity to changes in subsurface density. To further assess the reliability of our DL-based method, we change our target from density to saturation. We train our DL-based inversion using 2D surface gravity maps (500m sensor resolution) as input and 3D subsurface saturation as our target. The results of this inversion are shown in Table \ref{tab:results-saturation}, where we see that our DL-based approach yields similar results to inversion with changes in density as the target. This similar performance shows that DL-based inversion can map changes in surface gravity to multiple subsurface physical property models.

\setlength\tabcolsep{10pt}
\begin{table}[ht!]
\centering
\bgroup
\def\arraystretch{1.25}%  1 is the default, change whatever you need
\begin{tabular}{lcccc}
\hline
\multirow{2}{*}{} & \multicolumn{1}{c}{\multirow{2}{*}{Mean $\pm$ Std}} & \multicolumn{1}{l}{\multirow{2}{*}{Median}} & \multicolumn{2}{c}{Percentile} \\ \cline{4-5} 
                & \multicolumn{1}{l}{} & \multicolumn{1}{l}{} & 25$^\text{th}$ & 75$^\text{th}$ \\ \hline
MSE (su)  & 6.3e-05 $\pm$ 3.3e-05 & 7.4e-05 & 3.3e-05        & 8.5e-05          \\ \hline
R-Squared       & 0.70 $\pm$ 0.16       & 0.76    & 0.68           & 0.80          \\ \hline
Dice            & 0.59 $\pm$ 0.05       & 0.59    & 0.56           & 0.62          \\ \hline
\end{tabular}
\egroup
\caption{Accuracy of DL-based inversion of surface gravity data with 3D saturation models as the target. We see that our DL-based approach yields similar results to inversion with changes in density as the target. This similar performance shows that DL-based inversion can map changes in surface gravity to multiple subsurface physical property models.}
\label{tab:results-saturation}
\end{table}

\section{Discussion}
\label{sec:diss}
The above introduced results indicate that our DL-based method provides an effective approach for the inversion of surface gravity data. Particularly, the DL-driven approach successfully detects and recovers a variety of realistic-looking plumes that vary significantly in shape and exhibit non-trivial density distributions. Our DL-driven inversion demonstrates improvements over the widely used conventional L2 inversion in the following aspects: Our custom U-Net successfully inverts surface gravity data in a completely unconstrained environment. Because L2 inversion is constrained to only update cells within the reservoir mask, it is unsuitable for detecting out-of-reservoir leaks, a key element to be considered when monitoring CO$_2$ storage sites. Our DL inversion also outperforms conventional inversion in terms of model misfit and recovering the plume geometry. When combined with L2 inversion, our DL method matches L2 inversion in terms of data misfit while improving model misfit. 

The DL approach consistently predicts CO$_2$ plume models across multiple sensor resolution grids. This robustness to coarser grid data can yield significant cost savings when designing sensor grids to monitor CO$_2$ storage sites. Additionally, our results demonstrate that DL-based inversion can successfully learn mappings to multiple geophysical property models such as saturation with little to no modification to the underlying method. 

The benefits of DL-based inversion are limited from a computational perspective because the conventional gravity inversion method is not computationally demanding compared with seismic inversion. L2 inversion does not require training data and does not require high computational capacity, such as what GPUs provide. Training our DL-based inversion to convergence takes roughly $3$ hours when training on $4$ NVIDIA A100 GPUs, notice here that our U-Net implementation is not by any means computationally optimized. However, at test time, our DL-based method is comparable in speed to L2 inversion, taking less than one second to produce multiple predictions. 

The surface gravity data does not carry explicit plume depth information, limiting the information available in the input to the U-Net model. A pseudo-depth representation created from gravity data may improve our approach. Araya-Polo's work (published later by Chen et al.) obtained depth information via a downward continuation of gravity data that transforms 2D surface maps into 3D data, resulting in accurate segmentation results of salt structures from surface gravity data \cite{continuation}. Liu et al. added a pseudo-depth channel to the 2D resistivity data, which improved model resolution \cite{depth-weighting}. These works suggest that our results may benefit from including depth information in our DL-driven approach. 

Jointly inverting gravity data with other modalities like electromagnetic, resistivity, or seismic imaging may also improve the performance of our DL-driven approach. Yen et al. used a combination of seismic and electromagnetic data to recover salt bodies successfully \cite{joint-inversion}. In their approach, the joint inversion of seismic and electromagnetic data outperformed inversions with just each of the individual modalities. For CO$_2$ monitoring, Um et al. use joint inversion with surface gravity data, seismic, and electromagnetic imaging to improve their results compared to inversion with gravity alone \cite{um2022}. These works suggest that joint inversion with other imaging modalities is a potential avenue for improving DL-based inversion. 
%\todo[inline]{wrong place, most of the above paragraph should go to the literature review, here only the 3D gravity representation as a potential improvement comment}

In terms of generalization, our training data comes from a single reservoir. As a result, our model does not generalize well to surface gravity readings from other sources, such as the \emph{Snohvit} field. To better generalize our model, we hypothesize that more data is needed from multiple locations. Generative adversarial networks (GANs) may also help extend our DL-based inversion's generalization power. Yen et al. use GANs to enrich their training data in \cite{gans-1} and improve the performance of their segmentation model for seismic salt segmentation. Transfer learning presents an additional mitigation strategy for the lack of diversity in a training dataset. Using a model trained on the Johansen formation, we can use the weights of this model as an initial guess to train a model on a much smaller training set from another source. El Zini et al. and Yen et al. use transfer learning in their work to improve the performance of their DL models and improve their generalization power \cite{transfer-1}, \cite{gans-1}. Transfer learning achieves faster convergence compared to random initialization of networks \cite{transfer-2}. This speed-up may also mitigate the computational bottlenecks associated with DL-based inversion by dramatically decreasing the time required to train a model.

A recent study by Colombo et al. suggests a practical hybrid DL-aided geophysical inversion approach that iteratively minimizes model and data misfit \cite{physics-informed-1}. This hybrid approach benefits from data-driven DL-inversion and physics-based inversion (i.e., least squares optimization). Physics-informed DL inversion is emerging as a viable approach for practical geophysical inversion \cite{physics-informed-2}. However, the similarity between our physics-driven workflow and pure DL-driven results indicates that surface gravity inversion may not benefit from a hybrid approach. One possible explanation for this similarity is that gravity inversion is an ill-posed problem. Good density model fit does not necessarily correspond to good data fit and vice versa. During training, our workflow can converge in model misfit but does not converge in our forward model loss. The computational cost of appending our forward gravity model to our DL model during training increases the time required for convergence. Without transfer learning, our physics-driven workflow takes roughly four days to converge. With transfer learning, that time decreases to eight hours. In either case, we see a considerable slow down compared to our pure DL-based approach, which takes about three hours to converge. Improving our proposed physics-driven workflow in terms of accuracy and computational cost is a topic of ongoing and future work. 

Our time-dependent results indicate that DL-based inversion can successfully incorporate time-series gravity data for monitoring sequestered CO$_2$. However, our results are only tested on a single simulation of the Johansen formation. Further testing of our time-dependent method on multiple simulations of the same formation is required before we can make rigorous conclusions about the accuracy of our proposed method. From a computational point of view, adding our LSTM network to the training pipeline does not significantly affect the time or memory required for training our network. 

As mentioned earlier, the benefits of DL-based inversion are limited from a computational perspective. DL-based inversion requires a large amount of training data (simulated, recorded, or both) and computing infrastructure that can support training DL models. In the absence of both, conventional inversion methods like constrained L2 inversion make more sense for practitioners. However, suppose users have the required data and computing infrastructure and are willing to pay the upfront costs of data generation and DL training. In that case, the benefits of DL-based inversion are clear. Our results indicate that our pure DL-driven and combined L2-DL approaches produce more accurate density model reconstructions from surface gravity data. While the results from our physics-driven workflow and time-dependent training are promising, further work is required to improve the computational cost (i.e., physics-driven workflow) or test the method's robustness (i.e., time-dependent training).

\section{Conclusions}
\label{sec:conclu}
We developed an effective DL-based inversion method to recover high-resolution, subsurface CO$_2$ density models from surface gravity. Our DL architecture is trained on realistic, physics-simulated CO$_2$ plumes and gravity data. This training approach mirrors real-world site data collection and practical gravity monitoring techniques. Our DL-based inversion outperforms traditional inversion techniques for our selected metrics, and our combined L2-DL approach provides additional benefits to our pure DL approach. While there is room for improvement, the results presented here are promising and represent, to the best of our knowledge, the first fully 3D DL-based inversion of surface gravity data derived from a physics simulation of a proposed CO$_2$ storage site. DL inversion facilitates near real-time monitoring of geologic carbon sequestration operations. It can provide site operators with prompt updates of subsurface CO$_2$ distributions to underpin risk management and mitigation decision-making. 

% \begin{itemize}
%     \item Computational costs compared to traditional methods
%     \item Is this more accurate than traditional methods?
%     \item What are limitations of this method for real world CO$_2$ monitoring?
% \end{itemize}

% \subsection{Comparison of Methods}
% \setlength\tabcolsep{10pt}
% \begin{table}[ht!]
% \centering
% \bgroup
% \def\arraystretch{1.25}%  1 is the default, change whatever you need
% \resizebox{\textwidth}{!}{%
% \begin{tabular}{lcc|cc|cc}
% \hline
% \multirow{2}{*}{} & \multicolumn{2}{c|}{L2 } & \multicolumn{2}{c|}{Deep Learning} & \multicolumn{2}{c}{L2-Deep Learning} \\ \cline{2-7} 
%                 & Mean $\pm$ Std   & Median & Mean $\pm$ Std  & Median & Mean $\pm$ Std  & Median \\ \hline
% MSE (kg/m$^3$)  & 0.67 $\pm$ 0.39  & 0.73   & 0.26 $\pm$ 0.15 & 0.27   & \textbf{0.13 $\pm$ 0.09} & \textbf{0.11}   \\
% MSE ($\mu$Gal)  & 0.00 $\pm$ 0.00  & 0.00   & 0.49 $\pm$ 0.86 & 0.13   & \textbf{0.00 $\pm$ 0.00} & \textbf{0.00}   \\
% R-Squared       & -0.25 $\pm$ 0.73  & -0.01   & 0.68 $\pm$ 0.17 & 0.73   & \textbf{0.89 $\pm$ 0.06} & \textbf{0.90}   \\
% Dice            & 0.44 $\pm$ 0.06  & 0.45   & \textbf{0.78 $\pm$ 0.03} & \textbf{0.79}   & 0.41 $\pm$ 0.09 & 0.43   \\ \hline
% \end{tabular}%
% }
% \egroup
% \caption{Comparison of metrics for L2 inversion, deep learning inversion, and combined L2-deep learning inversion. We see here that the combied L2-deep learning method outperforms the other methods with respect to all of the metrics with the exception of Dice.}
% \label{tab:compare-methods}
% \end{table}

\bibliographystyle{splncs04}
\bibliography{references}
\end{document}